\documentclass{article} 
\usepackage{iclr2017_conference,times}
\usepackage{amsmath}
\usepackage{amsfonts}
\usepackage{amssymb}
\usepackage{amsthm}
\usepackage{bm}
\usepackage{graphicx}
\usepackage{hyperref}
\usepackage{url}

\usepackage{xcolor}
\definecolor{darkgreen}{rgb}{0,0.6,0}

\title{Deep Information Propagation}

\author{Samuel S. Schoenholz\thanks{Work done as a member of the Google Brain Residency program (\url{g.co/brainresidency})} \\
Google Brain
\And Justin Gilmer$^*$\\
Google Brain
\And Surya Ganguli \\
Stanford University
\And Jascha Sohl-Dickstein\\
Google Brain} 

\iclrfinalcopy 

\begin{document}

\maketitle

\begin{abstract}
We study the behavior of untrained neural networks whose weights and biases are randomly distributed using mean field theory. We show the existence of depth scales that naturally limit the maximum depth of signal propagation through these random networks. Our main practical result is to show that random networks may be trained precisely when information can travel through them. Thus, the depth scales that we identify provide bounds on how deep a network may be trained for a specific choice of hyperparameters. As a corollary to this, we argue that in networks at the edge of chaos, one of these depth scales diverges. Thus arbitrarily deep networks may be trained only sufficiently close to criticality. We show that the presence of dropout destroys the order-to-chaos critical point and therefore strongly limits the maximum trainable depth for random networks. Finally, we develop a mean field theory for backpropagation and we show that the ordered and chaotic phases correspond to regions of vanishing and exploding gradient respectively.
\end{abstract}

\section{Introduction}

Deep neural network architectures have become ubiquitous in machine learning. The success of deep networks is due to the fact that they are highly expressive~\citep{montufar2014} while simultaneously being relatively easy to optimize~\citep{choromanska2015, goodfellow2014} with strong generalization properties~\citep{recht2015}. Consequently, developments in machine learning often accompany improvements in our ability to train increasingly deep networks. Despite this, designing novel network architectures is frequently equal parts art and science. This is, in part, because a general theory for neural networks that might inform design decisions has lagged behind the feverish pace of design. 

A pair of recent papers~\citep{poole2016,raghu2016} demonstrated that random neural networks are exponentially expressive in their depth. Central to their approach was the consideration of 
networks after random initialization, whose weights and biases were i.i.d. Gaussian distributed. In particular the paper by \citet{poole2016} developed a ``mean field'' formalism for treating wide, untrained, neural networks. They showed that these mean field networks exhibit an order-to-chaos transition as a function of the weight and bias variances. Notably the mean field formalism is not closely tied to a specific choice of activation function or loss.

In this paper, we demonstrate the existence of several characteristic ``depth'' scales that emerge naturally and control signal propagation in these random networks. We then show that one of these depth scales, $\xi_c$, diverges at the boundary between order and chaos. This result is insensitive to many architectural decisions (such as choice of activation function) and will generically be true at any order-to-chaos transition. We then extend these results to include dropout and we show that even small amounts of dropout destroys the order-to-chaos critical point and consequently removes the divergence in $\xi_c$. Together these results bound the depth to which signal may propagate through random neural networks. 

We then develop a corresponding mean field model for gradients and we show that a duality exists between the forward propagation of signals and the backpropagation of gradients. The ordered and chaotic phases that \citet{poole2016} identified correspond to regions of vanishing and exploding gradients, respectively. We demonstrate the validity of this mean field theory by computing gradients of random networks on MNIST. This provides a formal explanation of the `vanishing gradients' phenomenon that has long been observed in neural networks~\citep{bengio1993}. We continue to show that the covariance between two gradients is controlled by the same depth scale that limits correlated signal propagation in the forward direction.

Finally, we hypothesize that a necessary condition for a random neural network to be trainable is that information should be able to pass through it. Thus, the depth-scales identified here bound the set of hyperparameters that will lead to successful training. To test this ansatz we train ensembles of deep, fully connected, feed-forward neural networks of varying depth on MNIST and CIFAR10, with and without dropout. Our results confirm that neural networks are trainable precisely when their depth is not much larger than $\xi_c$. This result is dataset independent and is, therefore, a universal function of network architecture.

A corollary of these result is that asymptotically deep neural networks should be trainable provided they are initialized sufficiently close to the order-to-chaos transition. The notion of ``edge of chaos'' initialization has been explored previously. Such investigations have been both direct as in \citet{bertschinger2005,glorot2010} or indirect, through initialization schemes that favor deep signal propagation such as batch normalization~\citep{ioffe2015}, orthogonal matrix initialization~\citep{saxe2014}, random walk initialization~\citep{sussillo2014}, composition kernels~\citep{daniely2016}, or residual network architectures~\citep{he2015}. The novelty of the work presented here is two-fold. First, our framework predicts the depth at which networks may be trained even far from the order-to-chaos transition. While a skeptic might ask when it would be profitable to initialize a network far from criticality, we respond by noting that there are architectures (such as neural networks with dropout) where no critical point exists and so this more general framework is needed. Second, our work provides a formal, as opposed to intuitive, explanation for why very deep networks can only be trained near the edge of chaos.

\section{Background}

We begin by recapitulating the mean-field formalism developed in \citet{poole2016}. Consider a fully-connected, untrained, feed-forward, neural network of depth $L$ with layer width $N_l$ and some nonlinearity $\phi:\mathbb R\to\mathbb R$. Since this is an untrained neural network we suppose that its weights and biases are respectively i.i.d. as $W_{ij}^l\sim N(0,\sigma_w^2/N_l)$ and $b_i^l\sim N(0,\sigma_b^2)$. Notationally we set $z^l_i$ to be the pre-activations of the $l$th layer and $y_i^{l+1}$ to be the activations of that layer. Finally, we take the input to the network to be $y_i^0 = x_i$. The propagation of a signal through the network is described by the pair of equations,
\begin{equation}\label{eq:background-propgation}
z_i^l = \sum_j W^l_{ij} y_j^l + b_i^l \hspace{4pc} y^{l+1}_i = \phi(z_i^l).
\end{equation}
Since the weights and biases are randomly distributed, these equations define a probability distribution on the activations and pre-activations over an ensemble of untrained neural networks. The ``mean-field'' approximation is then to replace $z_i^l$ by a Gaussian whose first two moments match those of $z_i^l$. For the remainder of the paper we will take the mean field approximation as given.

Consider first the evolution of a single input, $x_{i;a}$, as it evolves through the network (as quantified by $y_{i;a}^l$ and $z_{i;a}^l$). Since the weights and biases are independent with zero mean, the first two moments of the pre-activations in the same layer will be,
\begin{align}
\mathbb E[ z_{i;a}^l] = 0\hspace{4pc}\mathbb E[z_{i;a}^lz_{j;a}^l] = q_{aa}^l\delta_{ij}
\end{align}
where $\delta_{ij}$ is the Kronecker delta. Here $q_{aa}^l$ is the variance of the pre-activations in the $l$th layer due to an input $x_{i;a}$ and it is described by the recursion relation,
\begin{equation}\label{eq:background-diagonal}
q_{aa}^l = \sigma_w^2\int\mathcal Dz \phi^2\left(\sqrt{q_{aa}^{l-1}}z\right) + \sigma_b^2
\end{equation}
where $\int\mathcal Dz = \frac1{\sqrt{2\pi}}\int dze^{-\frac12z^2}$ is the measure for a standard Gaussian distribution. Together these equations completely describe the evolution of a single input through a mean field neural network. For any choice of $\sigma_w^2$ and $\sigma_b^2$ with bounded $\phi$, eq.~\ref{eq:background-diagonal} has a fixed point at $q^* = \lim_{l\to\infty} q_{aa}^l$.

The propagation of a pair of signals, $x^0_{i;a}$ and $x^0_{i;b}$, through this network can be understood similarly. Here the mean pre-activations are trivially the same as in the single-input case. The independence of the weights and biases implies that the covariance between different pre-activations in the same layer will be given by, $\mathbb E[z_{i;a}^lz_{j;b}^l] = q_{ab}^l\delta_{ij}$. The covariance, $q_{ab}^l$, will be given by the recurrence relation,
\begin{equation}\label{eq:background-off-diagonal}
q_{ab}^l = \sigma_w^2\int\mathcal Dz_1\mathcal Dz_2 \phi(u_1)\phi(u_2) + \sigma_b^2
\end{equation} 
where $u_1 = \sqrt{q_{aa}^{l-1}}z_1$ and $u_2=\sqrt{q_{bb}^{l-1}}\left(c_{ab}^{l-1} z_1 + \sqrt{1-(c^{l-1}_{ab})^2}z_2\right)$, with $c_{ab}^l = q_{ab}^l/\sqrt{q_{aa}^lq_{bb}^l}$, are Gaussian approximations to the pre-activations in the preceding layer with the correct covariance matrix. Moreover $c^l_{ab}$ is the correlation between the two inputs after $l$ layers.

\begin{figure}[h]
\begin{center}
\includegraphics[width=0.95\linewidth]{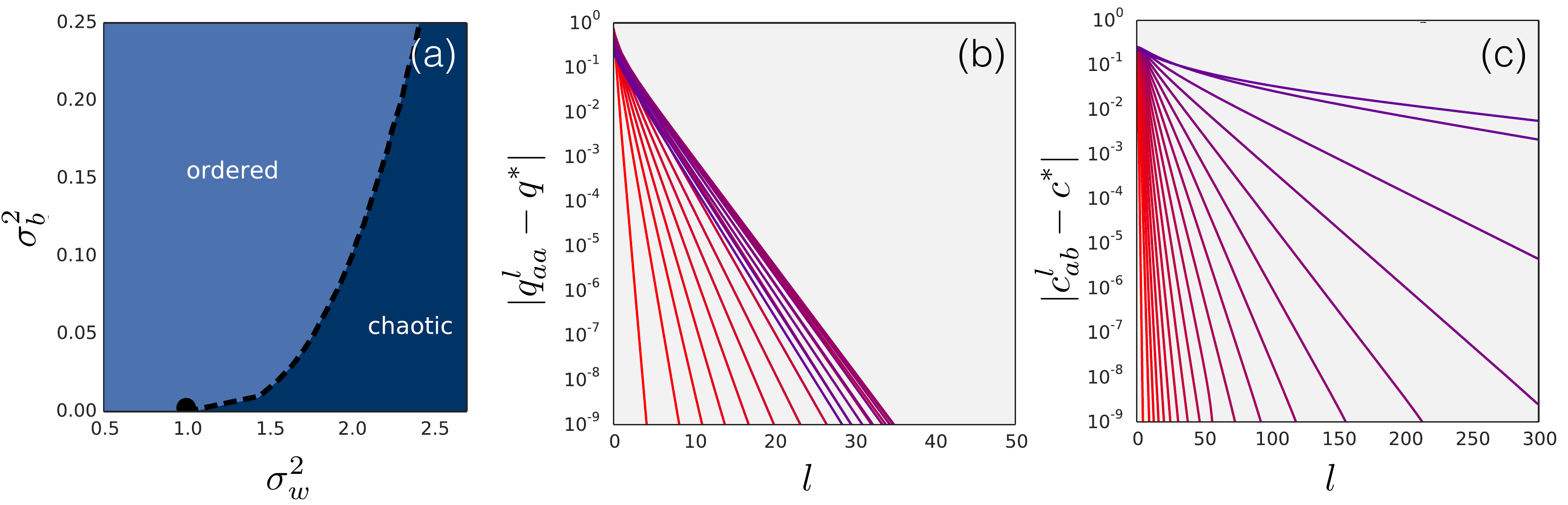}
\caption{Mean field criticality. (a) The mean field phase diagram showing the boundary between ordered and chaotic phases as a function of $\sigma_w^2$ and $\sigma_b^2$. (b) The residual $|q^* - q_{aa}^l|$ as a function of depth on a log-scale with $\sigma_b^2 = 0.05$ and $\sigma_w^2$ from 0.01 (red) to 1.7 (purple). Clear exponential behavior is observed. (c) The residual $|c^* - c_{ab}^l|$ as a function of depth on a log-scale. Again, the exponential behavior is clear. The same color scheme is used here as in (b). \label{fig:phase_diagram}}
\end{center}
\end{figure}

Examining eq.~\ref{eq:background-off-diagonal} it is clear that $c^* = 1$ is a fixed point of the recurrence relation. To determine whether or not the $c^* = 1$ is an attractive fixed point the quantity,
\begin{equation}
\chi_1 = \frac{\partial c_{ab}^l}{\partial c_{ab}^{l-1}} = \sigma_w^2\int\mathcal Dz\left[\phi'\left(\sqrt{q^*}z\right)\right]^2
\end{equation}  
is introduced. \citet{poole2016} note that the $c^* = 1$ fixed point is stable if $\chi_1 < 1$ and is unstable otherwise. Thus, $\chi_1 = 1$ represents a critical line separating an ordered phase (in which $c^* = 1$ and all inputs end up asymptotically correlated) and a chaotic phase (in which $c^*<1$ and all inputs end up asymptotically decorrelated). For the case of $\phi = \tanh$, the phase diagram in fig.~\ref{fig:phase_diagram} (a) is observed.

\section{Asymptotic Expansions and Depth Scales}

Our first contribution is to demonstrate the existence of two depth-scales that arise naturally within the framework of mean field neural networks. Motivating the existence of these depth-scales, we iterate eq.~\ref{eq:background-diagonal} and \ref{eq:background-off-diagonal} until convergence for many values of $\sigma_w^2$ between 0.1 and 3.0 and with $\sigma_b^2 = 0.05$ starting with $q_{aa}^0 = q_{bb}^0 = 0.8$ and $c_{ab}^0 = 0.6$. We see, in fig.~\ref{fig:phase_diagram} (b) and (c), that the manner in which both $q_{aa}^l$ approaches $q^*$ and $c_{ab}^l$ approaches $c^*$ is exponential over many orders of magnitude. We therefore anticipate that asymptotically $|q^l_{aa}-q^*| \sim e^{-l/\xi_q}$ and $|c^l_{ab}-c^*| \sim e^{-l/\xi_c}$ for sufficiently large $l$. Here, $\xi_q$ and $\xi_c$ define depth-scales over which information may propagate 
about the magnitude of a single input and the correlation between two inputs respectively.

We will presently prove that $q_{aa}^l$ and $c_{ab}^l$ are asymptotically exponential. In both cases we will use the same fundamental strategy wherein we expand one of the recurrence relations (either eq.~\ref{eq:background-diagonal} or eq.~\ref{eq:background-off-diagonal}) about its fixed point to get an approximate ``asymptotic'' recurrence relation. We find that this asymptotic recurrence relation in turn implies exponential decay towards the fixed point over a depth-scale, $\xi_{x}$. 

We first analyze eq.~\ref{eq:background-diagonal} and identify a depth-scale at which information about a single input may propagate. Let $q^l_{aa} = q^* + \epsilon^l$. By construction so long as $\lim_{l\to\infty}q^l_{aa} = q^*$ exists it follows that $\epsilon^l\to 0$ as 
$l\to\infty$. 
Eq.~\ref{eq:background-diagonal} may be expanded to lowest order in $\epsilon^l$ to arrive at an asymptotic recurrence relation (see Appendix \ref{sec:single-input}),
\begin{equation}\label{eq:asymptotic-diagonal}
\epsilon^{l+1} = \epsilon^l\left[\chi_1 + \sigma_w^2\int\mathcal Dz\phi''\left(\sqrt{q^*}z\right)\phi\left(\sqrt{q^*}z\right)\right] + \mathcal O\left((\epsilon^l)^2\right).
\end{equation}
Notably, the term multiplying $\epsilon^l$ is a constant. It follows that for large $l$ the asymptotic recurrence relation has an exponential solution, $\epsilon^l\sim e^{-l/\xi_q}$, with $\xi_q$ given by
\begin{equation}
\xi_q^{-1} = -\log\left[\chi_1 + \sigma_w^2\int\mathcal Dz\phi''\left(\sqrt{q^*}z\right)\phi\left(\sqrt{q^*}z\right)\right].
\end{equation}
This establishes $\xi_q$ as a depth scale that controls how deep information from a single input may penetrate into a random neural network.

Next, we consider eq.~\ref{eq:background-off-diagonal}. Using a similar argument (detailed in Appendix \ref{sec:double-input}) we can expand about $c^l_{ab} = c^* + \epsilon^l$ to find an asymptotic recurrence relation,
\begin{equation}
\epsilon^{l+1} = \epsilon^l\left[\sigma_w^2\int\mathcal Dz_1\mathcal Dz_2\phi'(u_1^*)\phi'(u_2^*)\right] + \mathcal O((\epsilon^l)^2).
\end{equation}
Here $u_1^* = \sqrt{q^*}z_1$ and $u_2^* = \sqrt{q^*}(c^*z_1 + \sqrt{1-(c^*)^2}z_2)$. Thus, once again, we expect that for large $l$ this recurrence will have an exponential solution, $\epsilon^l\sim e^{-l/\xi_c}$, with $\xi_c$ given by
\begin{equation}\label{eq:depth-scale-off-diagonal}
\xi_c^{-1} = -\log\left[\sigma_w^2\int\mathcal Dz_1\mathcal Dz_2\phi'(u_1^*)\phi'(u_2^*)\right].
\end{equation}
In the ordered phase $c^* = 1$ and so $\xi_c^{-1} = -\log \chi_1$. Since the transition between order and chaos occurs when $\chi_1 = 1$ it follows that $\xi_c$ diverges at any order-to-chaos transition so long as $q^*$ and $c^*$ exist.

\begin{figure}[h]
\begin{center}
\includegraphics[width=0.95\linewidth]{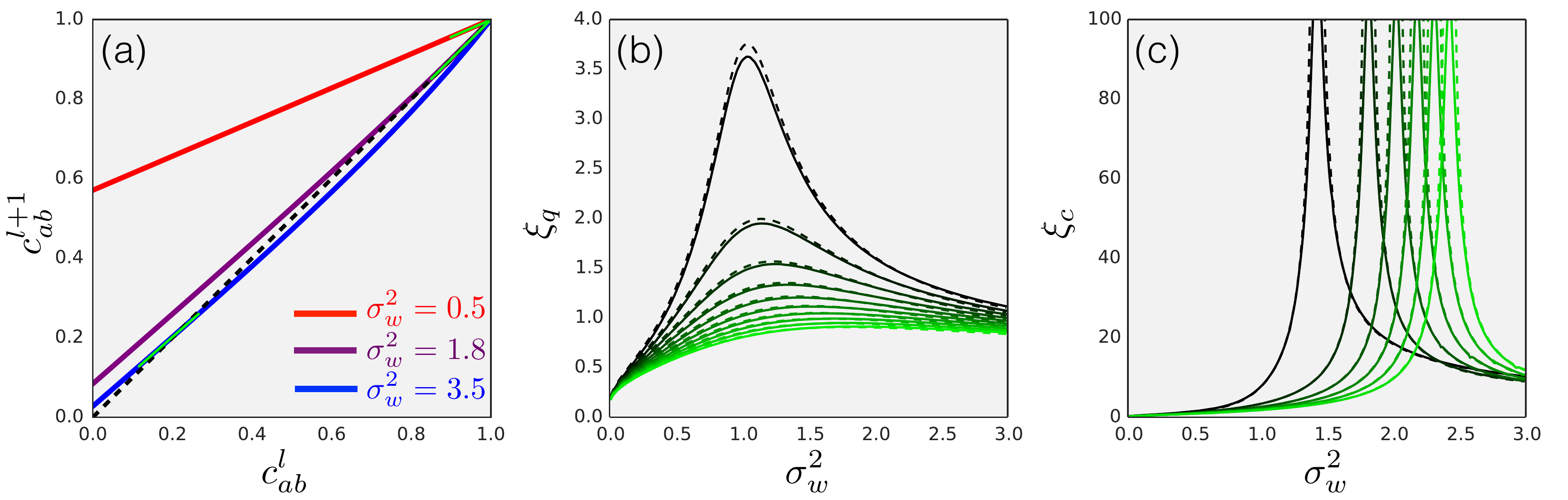}
\end{center}
\caption{Depth scales. (a) The iterative correlation map showing $c^{l+1}_{ab}$ as a function of $c^l_{ab}$ for three different values of $\sigma_w^2$. Green inset lines show the linearization of the iterative map about the critical point, $e^{-1/\xi_c}$. The three curves show networks far in the ordered regime (red), at the edge of chaos (purple), and deep in the chaotic regime (blue). (b) The depth scale for information propagated in a single input, $\xi_q$ as a function of $\sigma_w^2$ for $\sigma_b^2 = 0.01$ (black) to $\sigma_b^2 = 0.3$ (green). Dashed lines show theoretical predictions while solid lines show measurements. (c) The depth scale for correlations between inputs, $\xi_c$ for the same values of $\sigma_b^2$. Again dashed lines are the theoretical predictions while solid lines show measurements. Here a clear divergence is observed at the order-to-chaos transition.\label{fig:depth_scale_comparison}}
\end{figure}

These results can be investigated intuitively by plotting $c^{l+1}_{ab}$ vs $c^l_{ab}$ in fig.~\ref{fig:depth_scale_comparison} (a). In the ordered phase there is only a single fixed point, $c^l_{ab} = 1$. In the chaotic regime we see that a second fixed point develops and the $c^l_{ab} = 1$ point becomes unstable. We see that the linearization about the fixed points becomes significantly closer to the trivial map near the order-to-chaos transition.

To test these claims we measure $\xi_q$ and $\xi_c$ directly by iterating the recurrence relations for $q_{aa}^l$ and $c_{ab}^l$ as before with $q_{aa}^0 = q_{bb}^0 = 0.8$ and $c_{ab}^0 = 0.6$. In this case we consider values of $\sigma_w^2$ between $0.1$ and $3.0$ and $\sigma_b^2$ between $0.01$ and $0.3$. For each hyperparameter settings we fit the resulting residuals, $|q_{aa}^l-q^*|$ and $|c_{ab}^l-c^*|$, to exponential functions and infer the depth-scale. We then compare this measured depth-scale to that predicted by the asymptotic expansion. The result of this measurement is shown in fig.~\ref{fig:depth_scale_comparison}. In general we see that the agreement is quite good. As expected we see that $\xi_c$ diverges at the critical point. 

As observed in \citet{poole2016} we see that the depth scale for the propagation of information in a single input, $\xi_q$, is consistently finite and significantly shorter than $\xi_c$. To understand why this is the case consider eq.~\ref{eq:asymptotic-diagonal} and note that for $\tanh$ nonlinearities the second term is always negative. Thus, even as $\chi_1$ approaches 1 we expect $\chi_1 + \sigma_w^2\int\mathcal Dz\phi''(\sqrt{q^*}z)\phi(\sqrt{q^*}z)$ to be substantially smaller than 1.

\subsection{Dropout}

The mean field formalism can be extended to include dropout. The main contribution here will be to argue that even infinitesimal amounts of dropout destroys the mean field critical point, and therefore limits the trainable network depth. In the presence of dropout the propagation equation, eq.~\ref{eq:background-propgation}, becomes,
\begin{equation}
z_i^l = \frac1\rho\sum_jW^l_{ij}p_j^ly_j^l + b_i^l
\end{equation}
where $p_j\sim\text{Bernoulli}(\rho)$ and $\rho$ is the dropout rate. As is typically the case we have re-scaled the sum by $\rho^{-1}$ so that the mean of the pre-activation is invariant with respect to our choice of dropout rate. 

Following a similar procedure to the original mean field calculation consider the fate of two inputs, $x^0_{i;a}$ and $x^0_{i;b}$, as they are propagated through such a random network. We take the dropout masks to be chosen independently for the two inputs mimicking the manner in which dropout is employed in practice. 
With dropout the diagonal term in the covariance matrix will be (see Appendix \ref{sec:dropout-variance}),
\begin{equation}\label{eq:dropout-diagonal}
\bar q_{aa}^l = \frac{\sigma_w^2}{\rho}\int\mathcal Dz\phi^2\left(\sqrt{\bar q_{aa}^{l-1}}z\right) + \sigma_b^2.
\end{equation}
The variance of a single input with dropout will therefore propagate in an identical fashion to the vanilla case with a re-scaling $\sigma_w^2 \to \sigma_w^2/\rho$. Intuitively, this result implies that, for the case of a single input, the presence of dropout simply increases the effective variance of the weights.

Computing the off-diagonal term of the covariance matrix similarly (see Appendix \ref{sec:dropout-covariance}),
\begin{equation}\label{eq:dropout-off-diagonal}
\bar q_{ab}^l = \sigma_w^2\int\mathcal Dz_1\mathcal Dz_2\phi(\bar u_1)\phi(\bar u_2) + \sigma_b^2
\end{equation}
with $\bar u_1$, $\bar u_2$, and $\bar c_{ab}^l$ defined by analogy to the mean field equations without dropout. Here, unlike in the case of a single input, the recurrence relation is identical to the recurrence relation without dropout. To see that $\bar c^* = 1$ is no longer a fixed point of these dynamics consider what happens to eq.~\ref{eq:dropout-off-diagonal} when we input $\bar c^l = 1$. For simplicity, we leverage the short range of $\xi_q$ to replace $\bar q_{aa}^l = \bar q_{bb}^l = \bar q^*$. We find (see Appendix~\ref{sec:dropout-fixedpoint}),
\begin{equation}
\bar c^{l+1}_{ab} = 1 - \frac{1-\rho}{\rho \bar q^*}\sigma_w^2\int\mathcal Dz\phi^2\left(\sqrt{\bar q^*}z\right).
\end{equation}
The second term is positive for any $\rho<1$. This implies that if $\bar c^l_{ab} = 1$ for any $l$ then $\bar c^{l+1}_{ab}< 1$. Thus, $c^* = 1$ is not a fixed point of eq.~\ref{eq:dropout-off-diagonal} for any $\rho < 1$. Since eq.~\ref{eq:dropout-off-diagonal} is identical in form to eq.~\ref{eq:background-off-diagonal} it follows that the depth scale for signal propagation with dropout will likewise be given by eq.~\ref{eq:depth-scale-off-diagonal} with the substitutions $q^*\to\bar q^*$ and $c^*\to\bar c^*$ computed using eq.~\ref{eq:dropout-diagonal} and eq.~\ref{eq:dropout-off-diagonal} respectively. Importantly, since there is no longer a sharp critical point with dropout we do not expect a diverging depth scale.

\begin{figure}[h]
\begin{center}
\includegraphics[width=0.95\linewidth]{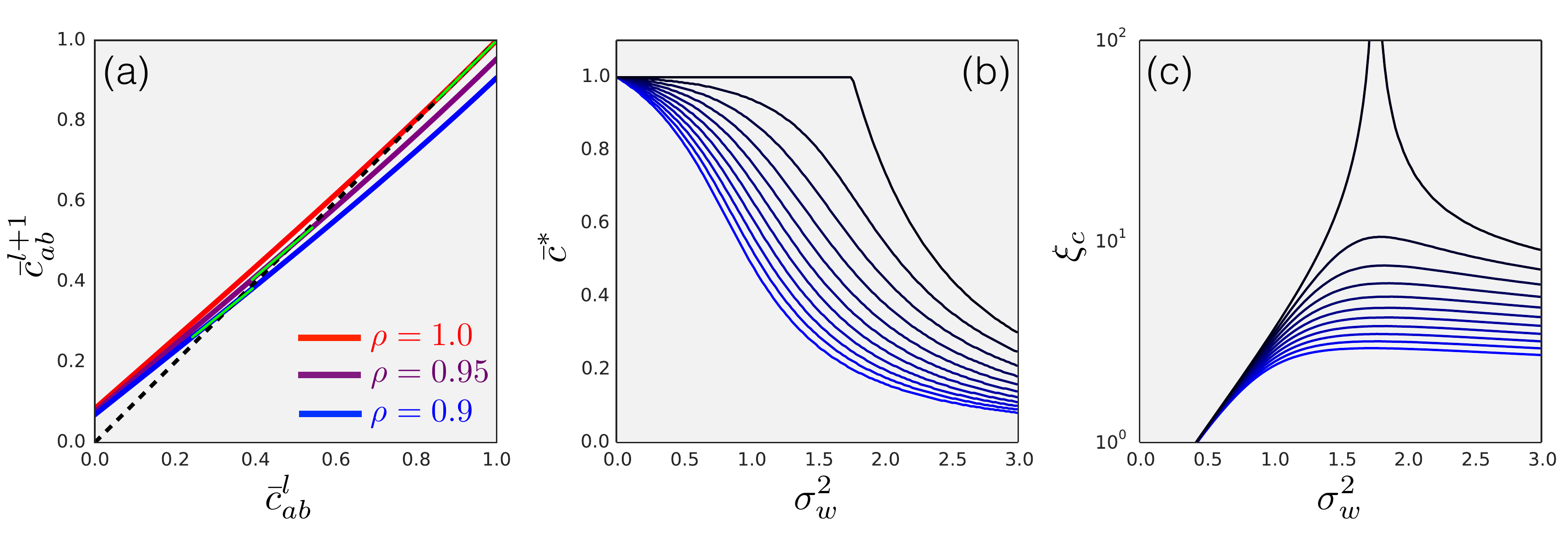}
\end{center}
\caption{Dropout destroys the critical point, and limits the depth to which information can propagate in a deep network. (a) The iterative correlation map showing $\bar c^{l+1}_{ab}$ as a function of $\bar c^l_{ab}$ for three different values of the dropout rate $\rho$ for networks tuned close to their critical point. Green inset lines show the linearization of the iterative map about the critical point, $e^{-1/\xi_c}$. (b) The asymptotic value of the correlation map, $c^*$, as a function of $\sigma_w^2$ for different values of dropout from $\rho = 1$ (black) to $\rho = 0.8$ (blue). We see that for all values of dropout except for $\rho = 1$, $c^*$ does not show a sharp transition between an ordered phase and a chaotic phase. (c) The correlation depth scale $\xi_c$ as a function of $\sigma_w^2$ for the same values of dropout as in (b). We see here that for all values of $\rho$ except for $\rho=1$ there is no divergence in $\xi_c$. \label{fig:dropout}}
\end{figure}

As in networks without dropout we plot, in fig.~\ref{fig:dropout} (a), the iterative map $\bar c^{l+1}_{ab}$ as a function of $\bar c^l_{ab}$. Most significantly, we see that the $\bar c^l_{ab} = 1$ is no longer a fixed point of the dynamics. Instead, as the dropout rate increases $\bar c^l_{ab}$ gets mapped to decreasing values and the fixed point monotonically decreases.

To test these results we plot in fig.~\ref{fig:dropout} (b) the asymptotic correlation, $c^*$, as a function of $\sigma_w^2$ for different values of dropout from $\rho = 0.8$ to $\rho=1.0$. As expected, we see that for all $\rho < 1$ there is no sharp transition between $c^* = 1$ and $c^* < 1$. Moreover as the dropout rate increases the correlation $c^*$ monotonically decreases. Intuitively this makes sense. Identical inputs passed through two different dropout masks will become increasingly dissimilar as the dropout rate increases. In fig.~\ref{fig:dropout} (c) we show the depth scale, $\xi_c$, as a function of $\sigma_w^2$ for the same range of dropout probabilities. We find that, as predicted, the depth of signal propagation with dropout is drastically reduced and, importantly, there is no longer a divergence in $\xi_c$. Increasing the dropout rate continues to decrease the correlation depth for constant $\sigma_w^2$.

\section{Gradient Backpropagation}

There is a duality between the forward propagation of signals and the backpropagation of gradients. To elucidate this connection consider the backpropagation equations given a loss $E$,
\begin{equation}
\frac{\partial E}{\partial W_{ij}^l} = \delta_i^l\phi(z_j^{l-1})\hspace{5pc} \delta_i^l = \phi'(z_i^l)\sum_j\delta_j^{l+1}W_{ji}^{l+1}
\end{equation} 
with the identification $\delta^l_i = \partial E/\partial z_i^l$. Within mean field theory, it is clear that the scale of fluctuations of the gradient of weights in a layer will be proportional to $\mathbb E[(\delta^l_i)^2]$ (see appendix \ref{sec:meanfield-gradient}).
In contrast to the pre-activations in forward propagation (eq. \ref{eq:background-propgation}), 
the $\delta^l_i$ will typically not be Gaussian distributed even in the large layer width limit.

Nonetheless, we can work out a recurrence relation for the variance of the error, $\tilde q_{aa}^{\ l} = \mathbb E[(\delta^l_i)^2]$, leveraging the Gaussian ansatz on the pre-activations. In order to do this, however, we must first make an additional approximation that the weights used during forward propagation are drawn independently from the weights used in backpropagation. This approximation is similar in spirit to the vanilla mean field approximation and is reminiscent of work on feedback alignment~\citep{lillicrap2014}. With this in mind we arrive at the recurrence (see appendix \ref{sec:meanfield-backprop}),
\begin{equation}\label{eq:gradient_variance}
\tilde q_{aa}^{\ l} = \tilde q_{aa}^{\ l+1}\frac{N_{l+1}}{N_l}\chi_1.
\end{equation}
The presence of $\chi_1$ in the above equation should perhaps not be surprising. In \citet{poole2016} they show that $\chi_1$ is intimately related to the tangent space of a given layer in mean field neural networks. We note that the backpropagation recurrence features an explicit dependence on the ratio of widths of adjacent layers of the network, $N_{l+1}/N_l$. Here we will consider exclusively constant width networks where this factor is unity. For a discussion of the case of unequal layer widths see \citet{glorot2010}.

Since $\chi_1$ depends only on the asymptotic $q^*$ it follows that for constant width networks we expect eq.~\ref{eq:gradient_variance} to again have an exponential solution with,
\begin{equation}\label{eq:gradient_depthscale}
\tilde q_{aa}^{\ l} = \tilde q_{aa}^{\ L}e^{-(L-l)/\xi_{{}_\nabla}} \hspace{4pc} \xi_{{}_\nabla}^{-1} = -\log\chi_1. 
\end{equation}
Note that here $\xi_{{}_\nabla}^{-1} = -\log\chi_1$ both above and below the transition. It follows that $\xi_{{}_\nabla}$ can be both positive and negative. We conclude that there should be three distinct regimes for the gradients. 
\begin{enumerate}
\item In the ordered phase, $\chi_1 < 1$ and so $\xi_{{}_\nabla}>0$. We therefore expect gradients to vanish over a depth $|\xi_{{}_\nabla}|$.
\item At criticality, $\chi_1 \to 1$ and so $\xi_{{}_\nabla}\to\infty$. Here gradients should be stable regardless of depth.
\item In the chaotic phase, $\chi_1 > 1$ and so $\xi_{{}_\nabla}<0$. It follows that in this regime gradients should explode over a depth $|\xi_{{}_\nabla}|$.
\end{enumerate}
Intuitively these three regimes make sense. To see this, recall that perturbations to a weight in layer $l$ can alternatively be viewed as perturbations to the pre-activations in the same layer. In the ordered phase both the perturbed signal and the unperturbed signal will be asymptotically mapped to the same point and the derivative will be small. In the chaotic phase the perturbed and unperturbed signals will become asymptotically decorrelated and the gradient will be large.

\begin{figure}[h]
\begin{center}
\includegraphics[width=0.7\linewidth]{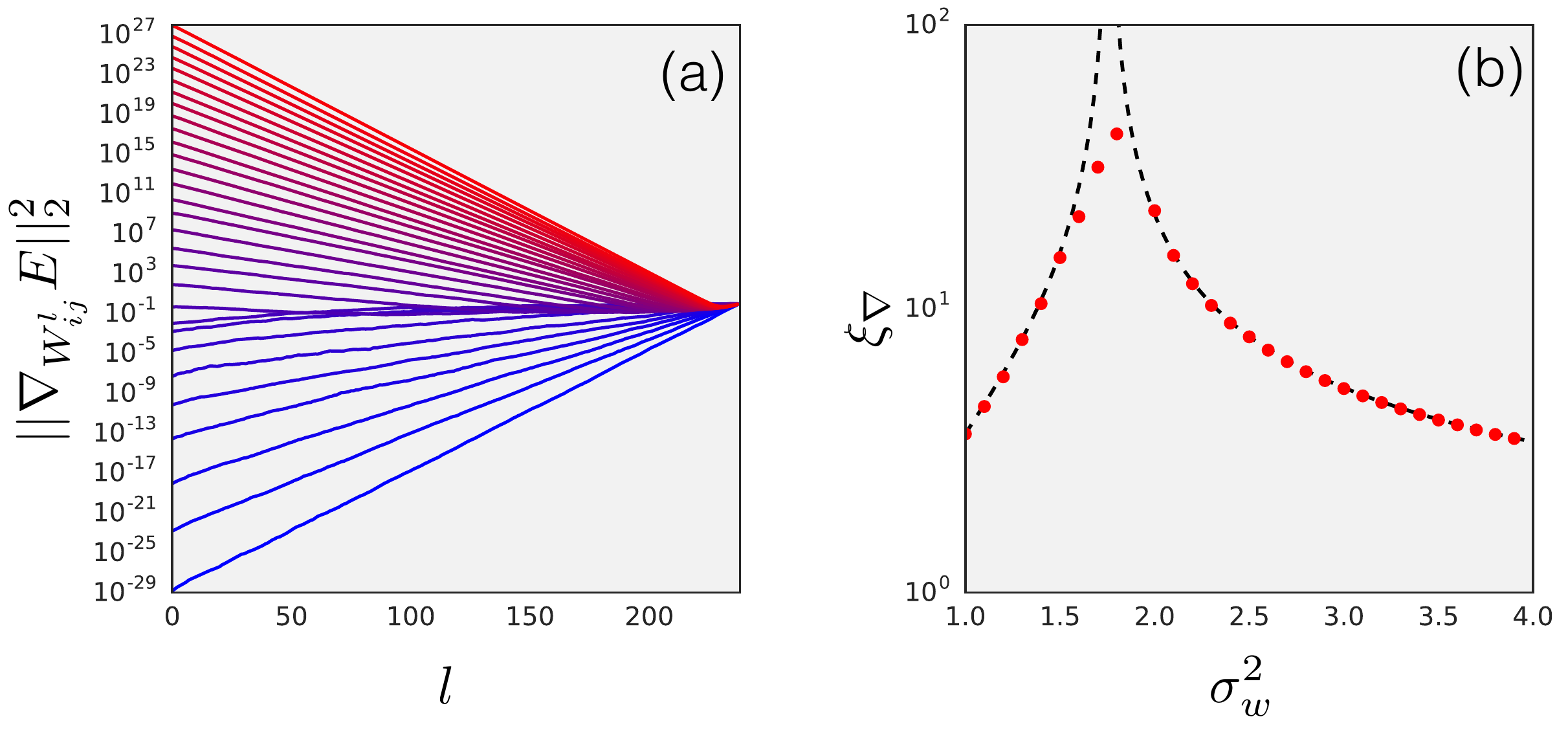}
\end{center}
\caption{Gradient backpropagation behaves similarly to signal forward propagation. (a) The 2-norm, $||\nabla_{W_{ab}^l}E||_2^2$ as a function of layer, $l$, for a 240 layer random network with a cross-entropy loss on MNIST. Different values of $\sigma_w^2$ from 1.0 (blue) to 4.0 (red) are shown. Clear exponential vanishing / explosion is observed over many orders of magnitude. (b) The depth scale for gradients predicted by theory (dashed line) compared with measurements from experiment (red dots). Similarity between theory and experiment is clear. Deviations near the critical point are primarily due to finite size effects.\label{fig:gradient}}
\end{figure}

To investigate these predictions we construct deep random networks of depth $L = 240$ and layer-width $N_l = 300$. We then consider the cross-entropy loss of these networks on MNIST. In fig.~\ref{fig:gradient} (a) we plot the layer-by-layer 2-norm of the gradient, $||\nabla_{W_{ab}^l}E||_2^2$, as a function of layer, $l$, for different values of $\sigma_w^2$. We see that $||\nabla_{W_{ab}^l}E||_2^2$ behaves exponentially over many orders of magnitude. Moreover, we see that the gradient vanishes in the ordered phase and explodes in the chaotic phase. We test the quantitative predictions of eq.~\ref{eq:gradient_depthscale} in fig.~\ref{fig:gradient} (b) where we compare $|\xi_{{}_\nabla}|$ as predicted from theory with the measured depth-scale constructed from exponential fits to the gradient data. Here we see good quantitative agreement between the theoretical predictions from mean field random networks and experimentally realized networks. Together these results suggest that the approximations on the backpropagation equations were representative of deep, wide, random networks.

Finally, we show that the depth scale for correlated signal propagation likewise controls the depth at which information stored in the covariance between gradients can survive. The existence of consistent gradients across similar samples from a training set ought to be especially important for determining whether or not a given neural network architecture can be trained. To establish this depth-scale first note (see Appendix~\ref{sec:meanfield-gradient-covariance}) that the covariance between gradients of two different inputs, $x_{i;1}$ and $x_{i;2}$, will be proportional to $(\nabla_{W_{ij}^l}E_a)\cdot(\nabla_{W_{ij}^l}E_b) \sim \mathbb E[\delta_{i;a}^l\delta_{i;b}^l] = \tilde q_{ab}^{\ l}$ where $E_a$ is the loss evaluated on $x_{i;a}$ and $\delta_{i;a} = \partial E_a/\partial z_{i;a}^l$ are appropriately defined errors. 

It can be shown (see Appendix~\ref{sec:meanfield-backprop-covariance}) that $\tilde q_{ab}^{\ l}$ features the recurrence relation,
\begin{equation}
\tilde q_{ab}^{\ l} = \tilde q_{ab}^{\ l+1}\frac{N_{l+1}}{N_{l+2}}\sigma_w^2\int\mathcal Dz_1\mathcal Dz_2\phi'(u_1)\phi'(u_2)
\end{equation}
where $u_1$ and $u_2$ are defined similarly as for the forward pass. Expanding asymptotically it is clear that to zeroth order in $\epsilon^l$, $\tilde q_{ab}^{l}$ will have an exponential solution with $\tilde q_{ab}^{\ l} = \tilde q_{ab}^{\ L}e^{-(L-l)/\xi_c}$ with $\xi_c$ as defined in the forward pass.

\section{Experimental Results}

Taken together, the results of this paper lead us to the following hypothesis: a necessary condition for a random network to be trained is that information about the inputs should be able to propagate forward through the network, and information about the gradients should be able to propagate backwards through the network. The preceding analysis shows that networks will have this property precisely when the network depth, $L$, is not much larger than the depth-scale $\xi_c$. This criterion is data independent and therefore offers a ``universal'' constraint on the hyperparameters that depends on network architecture alone. We now explore this relationship between depth of signal propagation and network trainability empirically.


\begin{figure}[h]
\begin{center}
\includegraphics[width=0.6\linewidth]{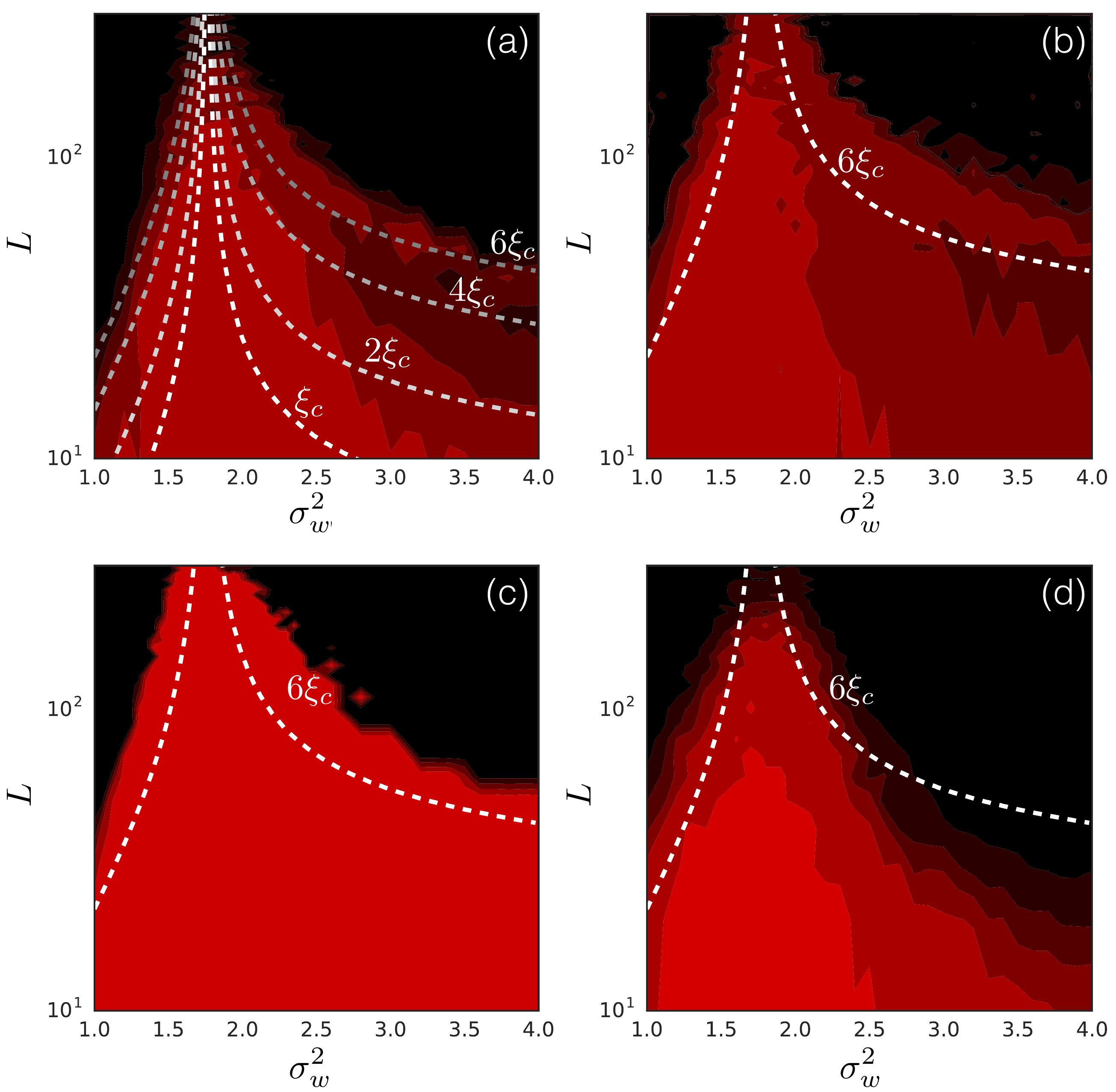}
\end{center}
\caption{Mean field depth scales control trainable hyperparameters. The training accuracy for neural networks as a function of their depth and initial weight variance, $\sigma_w^2$ from a high accuracy (red) to low accuracy (black). In (a) we plot the training accuracy after 200 training steps on MNIST using SGD. Here overlayed in grey dashed lines are different multiples of the depth scale for correlated signal propagation, $n\xi_c$.  We plot the accuracy in (b) after 2000 training steps on CIFAR10 using SGD, in (c) after 14000 training steps on MNIST using SGD, and in (d) after 300 training steps on MNIST using RMSPROP. Here we overlay in white dashed lines $6\xi_c$. \label{fig:experimental_no_dropout}}
\end{figure}

To investigate this prediction, we consider random networks of depth $10 \leq L \leq 300$ and $1 \leq \sigma_w^2 \leq 4$ with $\sigma_b^2 = 0.05$. We train these networks using Stochastic Gradient Descent (SGD) and RMSProp on MNIST and CIFAR10. We use a learning rate of $10^{-3}$ for SGD when $L\lesssim 200$, $10^{-4}$ for larger $L$, and $10^{-5}$ for RMSProp. These learning rates were selected by grid search between $10^{-6}$ and $10^{-2}$ in exponentially spaced steps of size $10$. We note that the depth dependence of learning rate was explored in detail in \citet{saxe2014}. In fig.~\ref{fig:experimental_no_dropout} (a)-(d) we color in red the training accuracy that neural networks achieved as a function of $\sigma_w^2$ and $L$ for different datasets, training time, and choice of minimizer (see Appendix~\ref{sec:further-experiments} for more comparisons). In all cases the neural networks over-fit the data to give a training accuracy of $100\%$ and test accuracies of $98\%$ on MNIST and $55\%$ on CIFAR10. We emphasize that the purpose of this study is to demonstrate trainability as opposed to optimizing test accuracy. 

We now make the connection between the depth scale, $\xi_c$, and the maximum trainable depth more precise. Given the arguments in the preceding sections we note that if $L = n\xi_c$ then signal through the network will be attenuated by a factor of $e^n$. To understand how much signal can be lost while still allowing for training, we overlay in fig.~\ref{fig:experimental_no_dropout} (a) curves corresponding to $n\xi_c$ from $n=1$ to $6$. We find that networks appear to be trainable when $L\lesssim 6\xi_c$. It would be interesting to understand why this is the case. 

Motivated by this argument in fig.~\ref{fig:experimental_no_dropout} (b)-(d) in white, dashed, overlay we plot twice the predicted depth scale, $6\xi_c$. There is clearly a relationship between the depth of correlated signal propagation and whether or not these networks are trainable. Networks closer to their critical point appear to train more quickly than those further away. Moreover, this relationship  has no obvious dependence on dataset, duration of training, or minimizer. We therefore conclude that these bounds on trainable hyperparameters are universal. This in turn implies that to train increasingly deep networks, one must generically be ever closer to criticality. 

\begin{figure}[h]
\begin{center}
\includegraphics[width=0.9\linewidth]{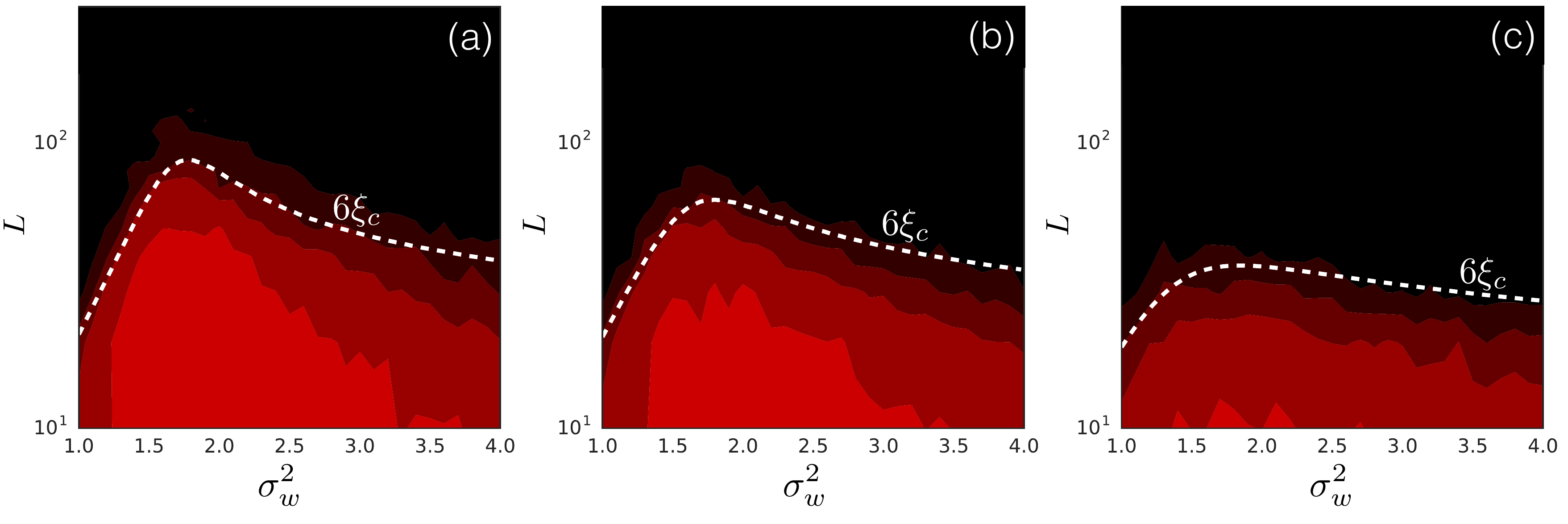}
\end{center}
\caption{The effect of dropout on trainability. The same scheme as in fig.~\ref{fig:experimental_no_dropout} but with dropout rates of (a) $\rho=0.99$, (b) $\rho=0.98$, and (c) $\rho=0.94$. Even for modest amounts of dropout we see an upper bound on the maximum trainable depth for neural networks. We continue to see good agreement between the prediction of our theory and our experimental training accuracy.\label{fig:experimental_dropout}}
\end{figure}

Next we consider the effect of dropout. As we showed earlier, even infinitesimal amounts of dropout disrupt the order-to-chaos phase transition and cause the depth scale to become finite. However, since the effect of a single dropout mask is to simply re-scale the weight variance by $\sigma_w^2\to\sigma_w^2/\rho$, the gradient magnitude will be stable near criticality, while the input and gradient correlations will not be. This therefore offers a unique opportunity to test whether the relevant depth-scale is $|1/\log\chi_1|$ or $\xi_c$.

In fig.~\ref{fig:experimental_dropout} we repeat the same experimental setup as above on MNIST with dropout rates $\rho = 0.99, 0.98, $ and 0.94. We observe, first and foremost, that even extremely modest amounts of dropout limit the maximum trainable depth to about $L = 100$. We additionally notice that the depth-scale, $\xi_c$, predicts the trainable region accurately for varying amounts of dropout. 

\section{Discussion}

In this paper we have elucidated the existence of several depth-scales that control signal propagation in random neural networks. Furthermore, we have shown that the degree to which a neural network can be trained depends crucially on its ability to propagate information about inputs and gradients through its full depth. At the transition between order and chaos, information stored in the correlation between inputs can propagate infinitely far through these random networks. This in turn implies that extremely deep neural networks may be trained sufficiently close to criticality. However, our contribution goes beyond advocating for hyperparameter selection that brings random networks to be nearly critical. Instead, we offer a general purpose framework that predicts, at the level of mean field theory, which hyperparameters should allow a network to be trained. This is especially relevant when analyzing schemes like dropout where there is no critical point and which therefore imply an upper bound on trainable network depth.

An alternative perspective as to why information stored in the covariance between inputs is crucial for training can be understood by appealing to the correspondence between infinitely wide Bayesian neural networks and Gaussian Processes \citep{neal2012}. In particular the covariance, $q_{ab}^l$, is intimately related to the kernel of the induced Gaussian Process. It follows that cases in which signal stored in the covariance between inputs may propagate through the network correspond precisely to situations in which the associated Gaussian Process is well defined.

Our work suggests that it may be fruitful to investigate pre-training schemes that attempt to perturb the weights of a neural network to favor information flow through the network. In principle this could be accomplished through a layer-by-layer local criterion for information flow or by selecting the mean and variance in schemes like batch normalization to maximize the covariance depth-scale.

These results suggest that theoretical work on random neural networks can be used to inform practical architectural decisions. However, there is still much work to be done. For instance, the framework developed here does not apply to unbounded activations, such as rectified linear units, where it can be shown that there are phases in which eq.~\ref{eq:background-diagonal} does not have a fixed point. Additionally, the analysis here applies directly only to fully connected feed-forward networks, and will need to be extended to architectures with structured weight matrices such as convolutional networks. 

We close by noting that in physics it has long been known that, through renormalization, the behavior of systems near critical points can control their behavior even far from the idealized critical case. We therefore make the somewhat bold hypothesis that a broad class of neural network topologies will be controlled by the fully-connected mean field critical point.

\subsubsection*{Acknowledgments}

We thank Ben Poole, Jeffrey Pennington, Maithra Raghu, and George Dahl for useful discussions. We are additionally grateful to RocketAI for introducing us to Temporally Recurrent Online Learning and two-dimensional time. 

\bibliography{iclr2017_conference}
\bibliographystyle{iclr2017_conference}

\section{Appendix}

Here we present derivations of results from throughout the paper. 



    
    

    
\subsection{Single input depth-scale}\label{sec:single-input}

\textbf{Result:}

Consider the recurrence relation for the variance of a single input,
\begin{equation}
q_{aa}^l = \sigma_w^2\int\mathcal Dz\phi^2\left(\sqrt{q_{aa}^{l-1}}z\right) + \sigma_b^2
\end{equation}
and a fixed point of the dynamics, $q^*$. $q^l_{aa}$ can be expanded about the fixed point to yield the asymptotic recurrence relation,
\begin{equation}
\epsilon^{l+1} = \epsilon^l\left[\chi_1 + \sigma_w^2\int\mathcal Dz\phi''\left(\sqrt{q^*}z\right)\phi\left(\sqrt{q^*}z\right)\right] + \mathcal O\left((\epsilon^l)^2\right).
\end{equation}

\textbf{Derivation:}

We begin by first expanding to order $\epsilon^l$,
\begin{align}
q^* + \epsilon^{l+1} &= \sigma_w^2\int\mathcal Dz\left[\phi\left(\sqrt{q^*+\epsilon^l}z\right)\right]^2  + \sigma_b^2\\
&\approx \sigma_w^2\int\mathcal Dz\left[\phi\left(\sqrt{q^*}z + \frac12\frac{\epsilon^lz}{\sqrt{q^*}}\right)\right]^2 + \sigma_b^2\\
&\approx \sigma_w^2\int\mathcal Dz\left[\phi\left(\sqrt{q^*}z\right) + \frac12\frac{\epsilon^lz}{\sqrt{q^*}}\phi'\left(\sqrt{q^*}z\right)\right]^2 + \sigma_b^2 + \mathcal O((\epsilon^l)^2)\\
&\approx \sigma_w^2\int\mathcal Dz\phi^2\left(\sqrt{q^*}z\right) + \sigma_b^2 + \epsilon^l\frac{\sigma_w^2}{\sqrt{q^*}}\int\mathcal Dz z\phi\left(\sqrt{q^*}z\right)\phi'\left(\sqrt{q^*}z\right) + \mathcal O((\epsilon^l)^2)\\
&\approx q^* + \epsilon^l\frac{\sigma_w^2}{\sqrt{q^*}}\int\mathcal Dz z\phi(\sqrt{q^*}z)\phi'\left(\sqrt{q^*}z\right) + \mathcal O((\epsilon^l)^2).
\end{align}
We therefore arrive at the approximate reccurence relation,
\begin{equation}
\epsilon^{l+1} =\epsilon^l\frac{\sigma_w^2}{\sqrt{q^*}}\int\mathcal Dz z\phi(\sqrt{q^*}z)\phi'\left(\sqrt{q^*}z\right) + \mathcal O((\epsilon^l)^2).
\end{equation}
Using the identity, $\int\mathcal Dz z f(z) = \int\mathcal Dz f'(z)$ we can rewrite this asymptotic recurrence relation as,
\begin{align}
\epsilon^{l+1} &= \epsilon^l\left[\sigma_w^2\int\mathcal Dz\left[\phi'\left(\sqrt{q^*}z\right)\right]^2 + \sigma_w^2\int\mathcal Dz\phi''\left(\sqrt{q^*}z\right)\phi\left(\sqrt{q^*}z\right)\right] + \mathcal O((\epsilon^l)^2)\\
&=\epsilon^l\left[\chi_1 + \sigma_w^2\int\mathcal Dz\phi''\left(\sqrt{q^*}z\right)\phi\left(\sqrt{q^*}z\right)\right] + \mathcal O((\epsilon^l)^2)
\end{align}
as required.

\subsection{Two input depth-scale}\label{sec:double-input}

\textbf{Result:} 

Consider the recurrence relation for the co-variance of two input,
\begin{equation}
q_{ab}^l = \sigma_w^2\int\mathcal Dz_1\mathcal Dz_2\phi(u_1)\phi(u_2) + \sigma_b^2,
\end{equation}
a correlation between the inputs, $c_{ab}^l = q_{ab}^l/\sqrt{q_{aa}^lq_{bb}^l}$, and a fixed point of the dynamics, $c^*$. $c^l_{ab}$ can be expanded about the fixed point to yield the asymptotic recurrence relation,
\begin{equation}
\epsilon^{l+1} = \epsilon^l\left[\sigma_w^2\int\mathcal Dz_1\mathcal Dz_2\phi'(u_1)\phi'(u_2)\right] + \mathcal O\left((\epsilon^l)^2\right).
\end{equation}

\textbf{Derivation:}

Since the relaxation of $q^l_{aa}$ and $q^l_{bb}$ to $q^*$ occurs much more quickly than the convergence of $q_{ab}^l$ we approximate $q^l_{aa} = q^l_{bb} = q^*$ as in \citet{poole2016}. We therefore consider the perturbation $q_{ab}^l/q^* = c_{ab}^l = c^* + \epsilon^l$. It follows that we may make the approximation,
\begin{align}
u_2^l &= \sqrt{q^*}\left(c_{ab}^l z_1 + \sqrt{1-(c_{ab}^l)^2} z_2\right)\\
&\approx \sqrt{q^*}\left(c^* z_1 + \sqrt{1-(c^*)^2 - 2c^*\epsilon^l}z_2\right) + \sqrt{q^*}\epsilon^l z_1 + \mathcal O(\epsilon^2)\\
\end{align}
We now consider the case where $c^* < 1$ and $c^* =1$ separately; we will later show that these two results agree with one another. First we consider the case where $c^* < 1$ in which case we may safely expand the above equation to get,
\begin{equation}
u_2^l = \sqrt{q^*}\left(c^* z_1 + \sqrt{1-(c^*)^2}z_2\right) + \sqrt{q^*}\epsilon^l\left( z_1 - \frac{c^*}{\sqrt{1-(c^*)^2}} z_2\right) + \mathcal O(\epsilon^2).
\end{equation}
This allows us to in turn approximate the recurrence relation,
\begin{align}
c_{ab}^{l+1} &= \frac{\sigma_w^2}{q^*}\int\mathcal Dz_1\mathcal Dz_2\phi(u_1^*)\phi(u_2^l) + \sigma_b^2\\
&\approx \frac{\sigma_w^2}{q^*}\int\mathcal Dz_1\mathcal Dz_2\phi(u_1^*)\left[\phi(u_2^*) + \sqrt{q^*}\epsilon^l\left( z_1 - \frac{c^*}{\sqrt{1-(c^*)^2}} z_2\right)\phi'(u_2^*)\right] + \sigma_b^2 + \mathcal O(\epsilon^2)\\
&=c^* + \frac{\sigma_w^2}{\sqrt{q^*}}\epsilon^l\int\mathcal Dz_1\mathcal Dz_2\left( z_1 - \frac{c^*}{\sqrt{1-(c^*)^2}} z_2\right)\phi(u_1^*)\phi'(u_2^*)\\
&=c^* + \frac{\sigma_w^2}{\sqrt{q^*}}\epsilon^l\left[\int\mathcal Dz_1\mathcal Dz_2z_1\phi(u_1^*)\phi'(u_2^*) - \frac{c^*}{\sqrt{1-(c^*)^2}}\int\mathcal Dz_1\mathcal Dz_2z_2\phi(u_1^*)\phi'(u_2^*)\right]\\
&=c^* + \sigma_w^2\epsilon^l\left[\int\mathcal Dz_1\mathcal Dz_2(\phi'(u_1^*)\phi'(u_2^*) + c^*\phi(u_1^*)\phi''(u_2^*)) - c^*\int\mathcal Dz_1\mathcal Dz_2\phi(u_1^*)\phi''(u_2^*)\right]\\
&=c^* + \sigma_w^2\epsilon^l\int\mathcal Dz_1\mathcal Dz_2\phi'(u_1^*)\phi'(u_2^*).
\end{align}
where $u_1^*$ and $u_2^*$ are appropriately defined asymptotic random variables. This leads to the asymptotic recurrence relation,
\begin{equation}
\epsilon^{l+1} = \sigma_w^2\epsilon^l\int\mathcal Dz_1\mathcal Dz_2\phi'(u_1^*)\phi'(u_2^*)
\end{equation}
as required.

We now consider the case where $c^* = 1$ and $c^l_{ab} = 1 - \epsilon^l$. In this case the expansion of $u_2^l$ will become,
\begin{equation}
u_2^l = \sqrt{q^*}z_1 + \sqrt{2q^*\epsilon^l}z_2 - \sqrt{q^*}\epsilon^l z_1 +  \mathcal O(\epsilon^{3/2})
\end{equation}
and so the lowest order correction is of order $\mathcal O(\sqrt{\epsilon^l})$ as opposed to $\mathcal O(\epsilon^l)$. As usual we now expand the recurrence relation, noting that $u_2^* = u_1^*$ is independent of $z_2$ when $c^* = 1$ to find,
\begin{align}
c_{ab}^{l+1} &= \frac{\sigma_w^2}{q^*}\int\mathcal Dz_1\mathcal Dz_2\phi(u_1^*)\phi(u_2^l) + \sigma_b^2\\
&\approx \frac{\sigma_w^2}{q^*}\int\mathcal Dz_1\mathcal Dz_2\phi(u_1^*)\left[\phi(u_2^*) + \left(\sqrt{2q^*\epsilon^l}z_2 - \sqrt{q^*}\epsilon^lz_1 \right)\phi'(u_2^*) + q^*\epsilon^l z_2^2\phi''(u_2^*)\right] + \sigma_b^2\\
&= c^* + \sigma_w^2\epsilon^l\int\mathcal D z \phi(\sqrt{q^*}z)\left[\phi''(\sqrt{q^*}z) - \frac1{\sqrt{q^*}}z\phi'(\sqrt{q^*}z)\right]\\
&= c^* + \sigma_w^2\epsilon^l\left[\int\mathcal D z\phi(\sqrt{q^*}z)\phi''(\sqrt{q^*}z) - \frac1{\sqrt{q^*}}\int\mathcal D zz\phi(\sqrt{q^*}z)\phi'(\sqrt{q^*}z)\right]\\
&= c^* - \sigma_w^2\epsilon^l\int\mathcal Dz\left[\phi'(\sqrt{q^*}z)\right]^2
\end{align} 
It follows that the asymptotic recurrence relation in this case will be,
\begin{equation}\label{eq:c1}
\epsilon^{l+1} = -\epsilon^l\sigma_w^2\int \mathcal Dz\left[\phi'(\sqrt{q^*}z)\right]^2 = -\epsilon^l\chi_1.
\end{equation}
where $\chi_1$ is the stability condition for the ordered phase. We note that although the approximations were somewhat different the asymptotic recurrence relation for $c^* < 1$ reduces eq.~\ref{eq:c1} result for $c^* = 1$. We may therefore use \ref{eq:background-off-diagonal} for all $c^*$.

\subsection{Variance of an input with dropout}\label{sec:dropout-variance}

\textbf{Result:}

In the presence of dropout with rate $\rho$, the variance of a single input as it is passed through the network is described by the recurrence relation,
\begin{equation}
\bar q_{aa}^l = \frac{\sigma_w^2}{\rho}\int\mathcal Dz \phi^2\left(\sqrt{\bar q^{l-1}_{aa}}z\right) + \sigma_b^2.
\end{equation}

\textbf{Derivation:}

Recall that the recurrence relation for the pre-activations is given by,
\begin{equation}
z_i^l = \frac1\rho\sum_j W_{ij}^l p_j^l y_j^l + b_i^l
\end{equation}
where $p_j^l\sim\text{Bernoulli}(\rho)$. It follows that the variance will be given by,
\begin{align}
\bar q^l_{aa} &= \mathbb E[(z_i^l)^2]\\
&= \frac1{\rho^2}\sum_j\mathbb E[(W_{ij}^l)^2]\mathbb E[(\rho_j^l)^2]\mathbb E[(y_j^l)^2] + \mathbb E[(b_i^l)^2]\\
&= \frac{\sigma_w^2}{\rho}\int\mathcal Dz \phi^2\left(\sqrt{\bar q^{l-1}_{aa}}z\right) + \sigma_b^2.
\end{align}
where we have used the fact that $\mathbb E[(p_j^l)^2] = \rho$.

\subsection{Covariance of two inputs with dropout}\label{sec:dropout-covariance}

\textbf{Result:} 

The co-variance between two signals, $z_{i;a}^l$ and $z_{i;b}^l$, with separate i.i.d. dropout masks $p_{i;a}^l$ and $p_{i;b}^l$ is given by,
\begin{equation}
\bar q_{ab}^l = \sigma_w^2\int\mathcal Dz_1\mathcal Dz_2\phi(\bar u_1)\phi(\bar u_2) + \sigma_b^2.
\end{equation}
where, in analogy to eq.~\ref{eq:background-off-diagonal}, $\bar u_1 = \sqrt{\bar q^l_{aa}}z_1$ and $\bar u_2 = \sqrt{\bar q^l_{bb}}\left(\bar c_{ab}^l z_1 + \sqrt{1-(\bar c_{ab}^l)^2}z_2\right)$.

\textbf{Derivation:}

Proceeding directly we find that,
\begin{align}
\mathbb E[z_{i;a}^lz_{i;b}^l] &= \frac1{\rho^2}\sum_j\mathbb E[(W_{ij}^l)^2]\mathbb E[p_{j;a}^l]\mathbb E[p_{j;b}^l]\mathbb E[y_{j;a}^ly_{j;b}^l] + \mathbb E[b_i^l]\\
&=\sigma_w^2\int\mathcal Dz_1\mathcal Dz_2\phi(\bar u_1)\phi(\bar u_2) + \sigma_b^2
\end{align}
where we have used the fact that $\mathbb E[p_{i;a}^l] = \mathbb E[p_{i;b}^l] = \rho.$ We have also used the same substitution for $\mathbb E[y_{j;a}^ly_{j;b}^l]$ used in the original mean field calculation with the appropriate substitution.

\subsection{The lack of a $c^* =1$ fixed point with dropout}\label{sec:dropout-fixedpoint}

\textbf{Result:}

If $c^l_{ab} = 1$ then it follows that,
\begin{equation}
\bar c^{l+1}_{ab} = 1 - \frac{1-\rho}{\rho \bar q^*}\sigma_w^2\int\mathcal Dz\phi^2\left(\sqrt{\bar q^*}z\right)
\end{equation}
subject to the approximation, $q_{aa}^l \approx q_{bb}^l \approx q^*$. This implies that $c^{l+1}_{ab} < 1.$

\textbf{Derivation:}

Plugging in $c^l_{ab} = 1$ with $q_{aa}^l \approx q_{bb}^l \approx q^*$ we find that $\bar u_1 = \bar u_2 = \sqrt{q^*}z_1$. It follows that,
\begin{align}
c_{ab}^{l+1} &= \frac{q_{ab}^{l+1}}{q^*}\\
&= \frac1{q^*}\left[\sigma_w^2\int\mathcal Dz\phi^2\left(\sqrt{q^*}z\right) + \sigma_b^2\right]\\
&= \frac1{q^*}\left[\sigma_w^2(1 - \rho^{-1} + \rho^{-1})\int\mathcal Dz\phi^2\left(\sqrt{q^*}z\right) + \sigma_b^2\right]\\
&=\frac1{q^*}\left[\frac{\sigma_w^2}\rho\int\mathcal Dz\phi^2\left(\sqrt{q^*}z\right) + \sigma_b^2\right]  + \frac{\sigma_w^2}{q^*}(1-\rho^{-1})\int\mathcal Dz\phi^2\left(\sqrt{q^*}z\right)\\
&= 1 - \frac{1-\rho}{\rho \bar q^*}\sigma_w^2\int\mathcal Dz\phi^2\left(\sqrt{\bar q^*}z\right)
\end{align}
as required. Here we have integrated out $z_2$ since nether $\bar u_1$ nor $\bar u_2$ depend on it.

\subsection{Mean field gradient scaling}\label{sec:meanfield-gradient}

\textbf{Result:}

In mean field theory the expected magnitude of the gradient $||\nabla_{W_{ij}^l}E||^2$ will be proportional to $\mathbb E[(\delta^l_i)^2]$.

\textbf{Derivation:}

We first note that since the $W_{ij}^l$ are i.i.d. it follows that,
\begin{align}
||\nabla_{W_{ij}^l}E||^2 &= \sum_{ij}\left(\frac{\partial E}{\partial W_{ij}^l}\right)^2\\
&\approx N_lN_{l+1}\mathbb E\left[\left(\frac{\partial E}{\partial W_{ij}^l}\right)^2\right]
\end{align}
where we have used the fact that the first line is related to the sample expectation over the different realizations of the $W_{ij}^l$ to approximate it by the analytic expectation in the second line. In mean field theory since the pre-activations in each layer are assumed to be i.i.d. Gaussian it follows that,
\begin{equation}
    \mathbb E\left[\left(\frac{\partial E}{\partial W_{ij}^l}\right)^2\right] = \mathbb E[(\delta_i^l)^2]\mathbb E[\phi^2(z_j^{l-1})]
\end{equation}
and the result follows.

\subsection{Mean field backpropagation}\label{sec:meanfield-backprop}

\textbf{Result:}

In mean field theory the recursion relation for the variance of the errors, $\tilde q^{\ l} = \mathbb E[(\delta^l_i)^2]$ is given by,
\begin{equation}
\tilde q^{\ l}_{aa}  = \tilde q^{\ l+1}_{aa}\frac{N_{l+1}}{N_{l+2}}\chi_1(q_{aa}^l).
\end{equation}

\textbf{Derivation:}

Computing the variance directly and using mean field approximation,
\begin{align}
\tilde q^{\ l}_{aa} = \mathbb E[(\delta_{i;a}^l)^2] &= \mathbb E[(\phi'(z_{i;a}^l))^2]\sum_j\mathbb E[(\delta^{l+1}_{j;a})^2]\mathbb E[(W_{ji}^{l+1})^2]\\
&=\mathbb E[(\phi'(z_{i;a}^l))^2]\frac{\sigma_w^2}{N_{l+1}}\sum_j\mathbb E[(\delta^{l+1}_{j;a})^2]\\
&=\mathbb E[(\phi'(z_{i;a}^l))^2]\frac{N_{l+1}}{N_{l+2}}\sigma_w^2\tilde q^{\ l+1}_{aa}\\
&=\sigma_w^2\tilde q^{\ l+1}_{aa}\frac{N_{l+1}}{N_{l+2}}\int\mathcal Dz \left[\phi'\left(\sqrt{q_{aa}^l}z\right)\right]^2\\
&\approx\tilde q^{\ l+1}_{aa}\frac{N_{l+1}}{N_{l+2}}\chi_1
\end{align}
as required. In the last step we have made the approximation that $q^l_{aa} \approx q^*$ since the depth scale for the variance is short ranged.

\subsection{Mean field gradient covariance scaling}\label{sec:meanfield-gradient-covariance}

\textbf{Result:}

In mean field theory we expect the covariance between the gradients of two different inputs to scale as,
\begin{equation}
(\nabla_{W_{ij}^l}E_a)\cdot(\nabla_{W_{ij}^l}E_b)\sim\mathbb E[\delta_{i;a}\delta_{i;b}].
\end{equation}

\textbf{Derivation:}

We proceed in a manner analogous to Appendix~\ref{sec:meanfield-gradient}. Note that in mean field theory since the weights are i.i.d. it follows that
\begin{align}
(\nabla_{W_{ij}^l}E_a)\cdot(\nabla_{W_{ij}^l}E_b) &= \sum_{ij}\frac{\partial E_a}{\partial W_{ij}^l}\frac{\partial E_b}{\partial W_{ij}^l}\\
&\approx N_l N_{l+1}\mathbb E\left[\frac{\partial E_a}{\partial W_{ij}^l}\frac{\partial E_b}{\partial W_{ij}^l}\right]
\end{align}
where, as before, the final term is approximating the sample expectation. Since the weights in the forward and backwards passes are chosen independently it follows that we can factor the expectation as,
\begin{equation}
\mathbb E\left[\frac{\partial E_a}{\partial W_{ij}^l}\frac{\partial E_b}{\partial W_{ij}^l}\right] = \mathbb E[\delta_{i;a}^l\delta_{i;b}^l]\mathbb E[\phi(z_{i;a}^l)\phi(z_{i;b}^l)]
\end{equation}
and the result follows.

\subsection{Mean field backpropagation of covariance}\label{sec:meanfield-backprop-covariance}

\textbf{Result:}

The covariance between the gradients due to two inputs scales as,
\begin{equation}
\tilde q_{ab}^{\ l} = \tilde q_{ab}^{\ l+1}\frac{N_{l+1}}{N_{l+2}}\sigma_w^2\int\mathcal Dz_1\mathcal Dz_2\phi'(u_1)\phi'(u_2)
\end{equation}
under backpropagation.

\textbf{Derivation}

As in the analogous derivation for the variance, we compute directly,
\begin{align}
\tilde q^{\ l}_{ab} = \mathbb E[\delta_{i;a}^l\delta_{i;b}^l] &= \mathbb E\left[\phi'(z_{i;a})\phi'(z_{i;b})\right]\sum_j\mathbb E[\delta^{l+1}_{j;a}\delta^{l+1}_{j;b}]\mathbb E[(W_{ji}^{l+1})^2]\\
&=\tilde q_{ab}^{\ l+1}\frac{N_{l+1}}{N_{l+2}}\sigma_w^2\int\mathcal Dz_1\mathcal Dz_2\phi'(u_1)\phi'(u_2)
\end{align}
as required.

\subsection{Further experimental results}\label{sec:further-experiments}

Here we include some more experimental figures that investigate the effects of training time, minimizer, and dataset more closely.
\begin{figure}[h]
\begin{center}
\includegraphics[width=0.8\linewidth]{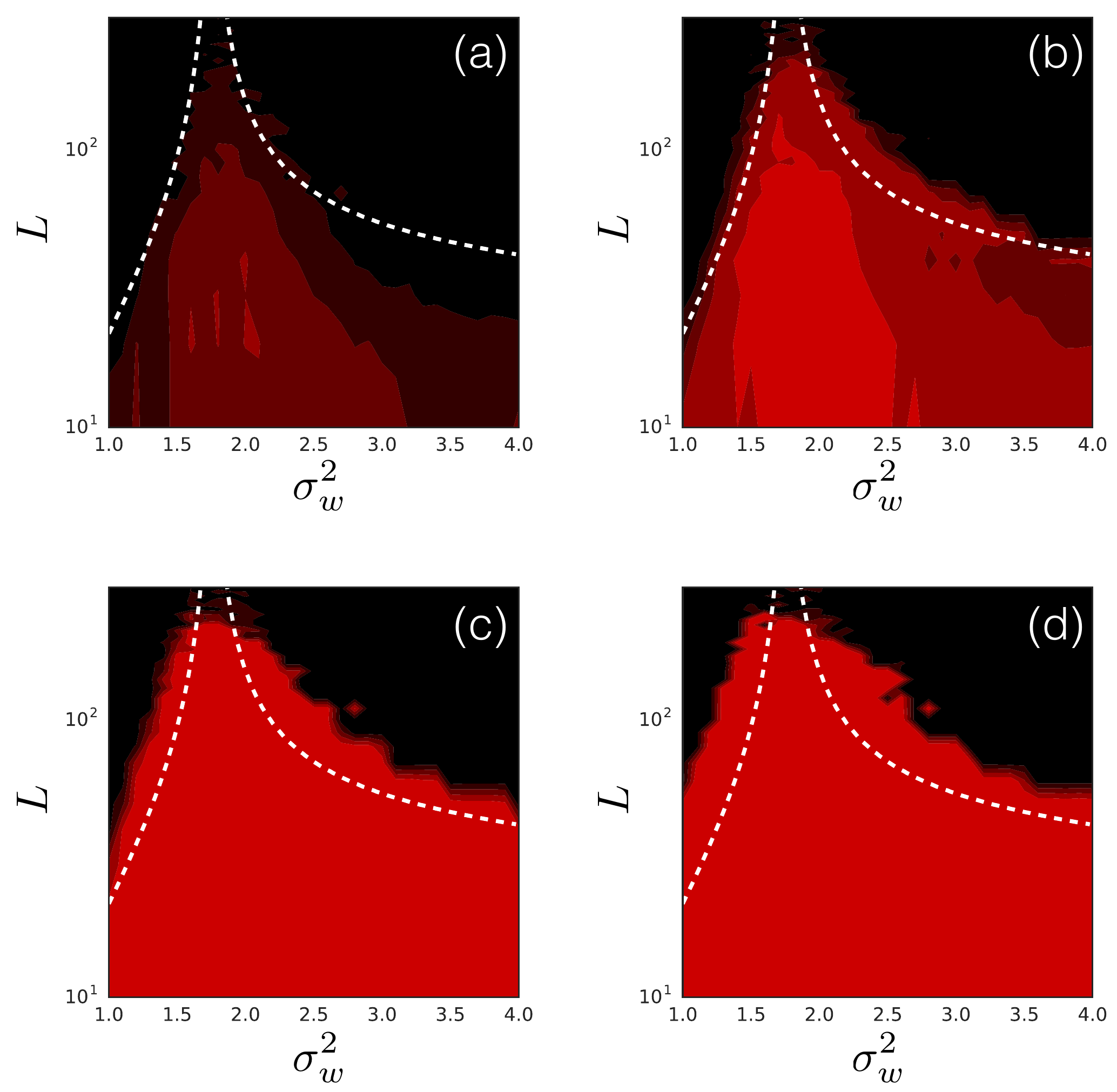}
\end{center}
\label{fig:experimental_no_dropout_a}
\caption{Training accuracy on MNIST after (a) 45 (b) 304 (c) 2048 and (d) 13780 steps of SGD with learning rate $10^{-3}$.}
\end{figure}

\begin{figure}[h]
\begin{center}
\includegraphics[width=0.8\linewidth]{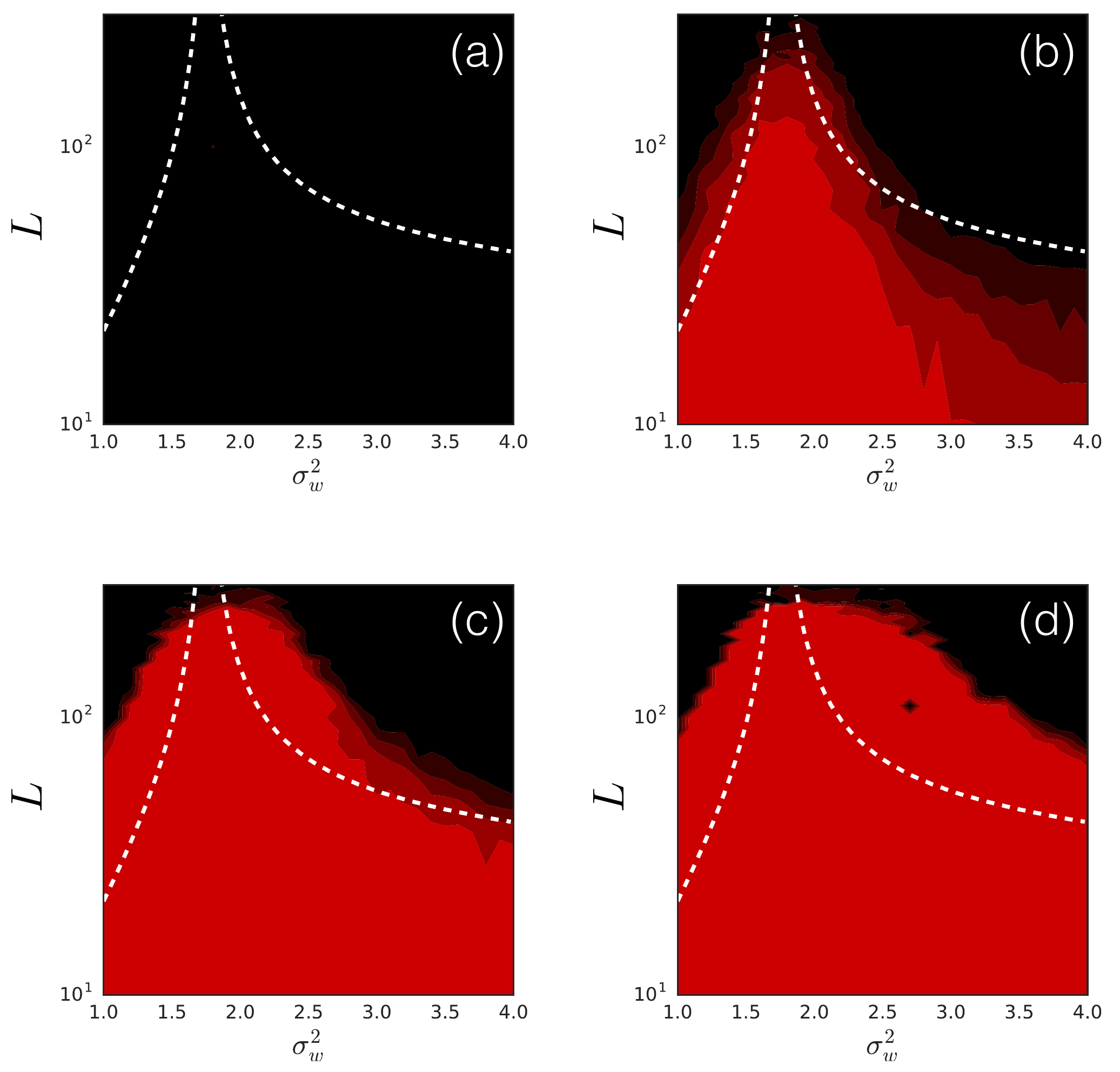}
\end{center}
\label{fig:experimental_no_dropout_b}
\caption{Training accuracy on MNIST after (a) 45 (b) 304 (c) 2048 and (d) 13780 steps of RMSProp with learning rate $10^{-5}$.}
\end{figure}

\end{document}